\documentclass[10pt,twocolumn,letterpaper]{article}

\usepackage{cvpr}              

\usepackage{graphicx}
\usepackage{amsmath}
\usepackage{amssymb}
\usepackage{booktabs}
\usepackage{multirow}
\usepackage{multirow}

\usepackage[misc]{ifsym}

\usepackage[pagebackref,breaklinks,colorlinks]{hyperref}

\usepackage[capitalize]{cleveref}
\crefname{section}{Sec.}{Secs.}
\Crefname{section}{Section}{Sections}
\Crefname{table}{Table}{Tables}
\crefname{table}{Tab.}{Tabs.}

\def\cvprPaperID{6336} 
\def\confName{CVPR}
\def\confYear{2022}

\begin{document}

\title{Learning Modal-Invariant and Temporal-Memory for Video-based Visible-Infrared Person Re-Identification}


\author{Xinyu Lin\textsuperscript{1}, Jinxing Li\textsuperscript{1 \Letter }, Zeyu Ma\textsuperscript{1}, Huafeng Li\textsuperscript{2}, Shuang Li\textsuperscript{2},Kaixiong Xu\textsuperscript{2}, Guangming Lu\textsuperscript{1}, David Zhang\textsuperscript{3}\\
\textsuperscript{1}Harbin Institute of Technology, Shenzhen, \textsuperscript{2}Kunming University of Science and Technology,\\ \textsuperscript{3}The Chinese University of HongKong, Shenzhen.\\
{\tt\small \{linxinyu0327, lijinxing158\}@gmail.com,} 
\tt\small zeyu.ma@stu.hit.edu.cn, \tt\small hfchina99@163.com, \\ 
\tt\small \{shuangli936, xukaixiong99\}@gmail.com, 
\tt\small luguangm@hit.edu.cn, 
\tt\small davidzhang@cuhk.edu.cn 
}


\maketitle
\begin{abstract}
	Thanks for the cross-modal retrieval techniques, visible-infrared (RGB-IR) person re-identification (Re-ID) is achieved by projecting them into a common space, allowing person Re-ID in 24-hour surveillance systems.
	However, with respect to the probe-to-gallery, almost all existing RGB-IR based cross-modal person Re-ID methods focus on image-to-image matching, while the video-to-video matching which contains much richer spatial- and temporal-information remains under-explored.
	In this paper, we primarily study the video-based cross-modal person Re-ID method.
	To achieve this task, a video-based RGB-IR dataset is constructed, in which 927 valid identities with 463,259 frames and 21,863 tracklets captured by 12 RGB/IR cameras are collected.
	Based on our constructed dataset, we prove that with the increase of frames in a tracklet, the performance does meet more enhancement, demonstrating the significance of video-to-video matching in RGB-IR person Re-ID. Additionally, a novel method is further proposed, which not only projects two modalities to a modal-invariant subspace, but also extracts the temporal-memory for motion-invariant. Thanks to these two strategies, much better results are achieved on our video-based cross-modal person Re-ID. The code and dataset are released at: {\url{https://github.com/VCM-project233/MITML}}.
	\footnote{\noindent \Letter \ Jinxing Li is the Corresponding Author.}
	
\end{abstract}


\section{Introduction}
\label{sec:intro}

Person re-identification (Re-ID)\cite{zheng2012reidentification,koestinger2012large,liao2015person,yang2014salient} focuses on matching probe pedestrian images with the gallery sets. Due to multiple views which are non-overlapped, there are significant changes in human body postures, illumination and backgrounds, leading a large challenge to Re-ID. Thanks to the rapid development of deep learning, various deep end-to-end approaches \cite{chen2017person,li2018harmonious,li2018unsupervised,li2019recover} have been studied, greatly enhancing the Re-ID performance.

\begin{figure}
	\centering
	\includegraphics[width=0.9\linewidth]{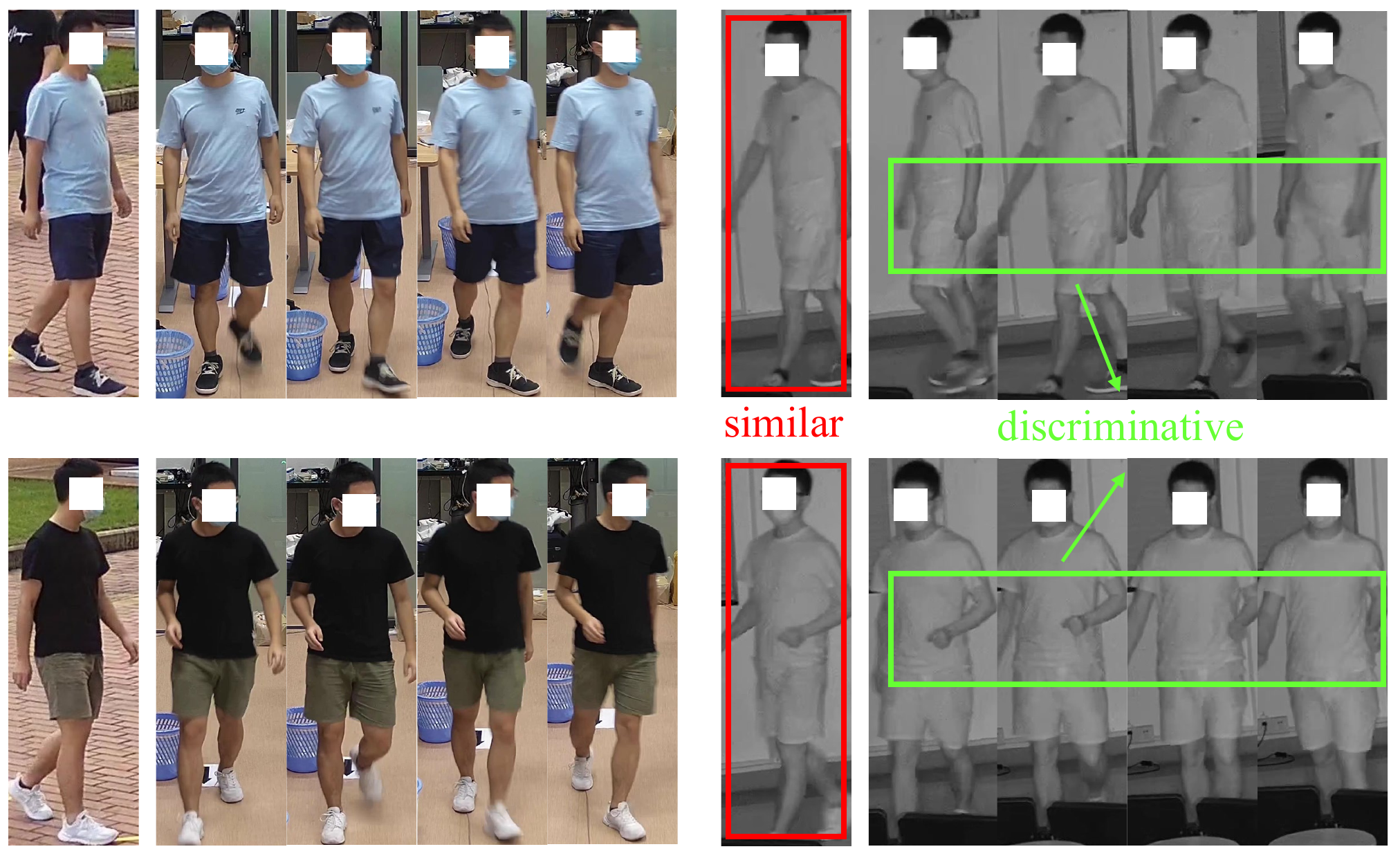}
	
	\caption{Advantages of video-based cross-modal person Re-ID. If two persons enjoy similar appearances, video data can also provide discriminative temporal-information that image data is unavailable. Specifically, the person wearing the black T-shirt is quite similar to the person wearing the blue T-shirt under the IR camera (shown in the red box), while their specific arm postures in the motions give the discriminative features (shown in the green box).}
	\label{fig:introduction}
\end{figure}

Despite the achievement of aforementioned methods, most of them are heavily dependent on the RGB images, so that the lighting for the cameras is essential. However, this constraint is too strict, especially at night, making the collected RGB data uninformative and failing to achieve person Re-ID. Fortunately, most surveillance cameras can automatically switch from RGB to the infrared (IR) mode if the lighting is unavailable.
In contrast to RGB images, IR images are capable of preserving the information under invisible lighting and showing pedestrians clearly.
Thus, in order to achieve person Re-ID in 24-hour surveillance systems, the RGB-IR based visible-infrared (cross-modal) person Re-ID \cite{wu2017rgb, dai2018cross,liao2015person} provides a promising strategy. For instance, Wu \textit{et al.}\cite{wu2017rgb} first collected an RGB-IR dataset and proved the feasibility for these two modalities matching. Inspired by this work, various cross-modal Re-ID works were then studied.


%

Although the cross-modal person Re-ID methods fill the gap between RGB images and IR images, they are only single-image based tasks. In the data collection, pedestrians originally appear in the video databases, containing multiple frames in each tracklet. Intuitively, the video-based data contains much richer visual information than a single image \cite{zheng2016mars}.
In some specific cases, it is indeed difficult to identify two persons with similar appearances if only a single image is given. This case is more difficult for the infrared modality, and even the human beings cannot guarantee the correctness. In contrast to still images, the video is an image sequence containing the spatial and temporal-information, so that the beneficial motion information can be exploited for the discriminative identification.
For instance, as displayed in \cref{fig:introduction}, two images captured from two persons enjoy similarity under the IR camera. However, the person wearing the black T-shirt has the specific arm posture in the motion, compared with the person wearing the blue T-shirt. Thanks to such motion characteristics, more discriminative information is provided for us to achieve a more robust and accurate identification model. Thus, it is quite significant to replace the still-images with videos in cross-modal person Re-ID.




To address this problem, in this paper, the video-based cross-modal Re-ID is studied.
In comparison to image-based cross-modal Re-ID, the video-based cross-modal Re-ID further aims to exploit the temporal-information for robust feature extraction. In contrast to the existing video-based RGB Re-ID methods, our focused work additionally extracts the consistency between RGB and IR modalities.


In order to achieve the video-based cross-modal Re-ID, an associated database is inevitable. Although Wu et al. \cite{wu2017rgb} has presented an RGB-IR dataset, it only focuses on the image-based retrieval, being far away from our video-based requirement. To substantiate our task, we primarily construct a video-based RGB-IR database named HITSZ Video Cross-Modal (HITSZ-VCM) Re-ID dataset.
The comparison between our collected dataset and existing Re-ID datasets is listed in \cref{tab:datasets}.
Different from SYSU-MM01\cite{wu2017rgb} which only collected the RGB images and IR images via 4 RGB cameras and 2 IR cameras, we set 12 cameras to capture both RGB and IR videos and much more valid identities are collected. 
Totally, 927 valid identities including 11,785 / 10,078 tracklets and 251,452 / 211,807 images with or free from masks for RGB and IR modalities are obtained, respectively.

For the video-based cross-modal Re-ID, the spatial- and temporal-information among each tracklet do contribute to the performance improvement.
In this paper, a baseline method is first applied to our constructed dataset, demonstrating the significance of video-based cross-modal Re-ID. Specifically, we follow Ye \textit{et al.}'s \cite{ye2020dynamic} baseline on image-based cross-modal Re-ID and add a module to utilize the temporal-information. Additionally, we also propose a novel method named Modal-Invariant and Temporal-Memory Learning (MITML). Two modalities are transformed to get modal-invariant but id-related features through an adversarial strategy, so that the gap between RGB and IR modalities is relieved. Referring to the motion information in a tracklet, we also propose a temporal memory refinement module to extract the temporal-information.
Thanks to these two strategies, the Re-ID performance on our dataset is further improved.

Overall, the main contributions of this paper are:
\begin{itemize}
	\item We construct a video-based RGB-IR database, allowing the study on video-based cross-modal person Re-ID. \textbf{Different from existing Re-ID works, to the best of our knowledge, this is the first work which jointly takes cross modalities and videos into account, defining a challenging task.}
	\item We introduce a baseline to prove the significance of video-based cross-modal person Re-ID. In detail, by embedding a temporal-information exploitation module, the cross-modal person Re-ID performance meets a continuous increase when the number of images in a tracklet rises.
	\item A novel method named Modal-Invariant and Temporal-Memory Learning (MITML) is additionally proposed by more efficiently removing modal-variance and exploiting motion information. Experimental results substantiate the superiority of our proposed method.
\end{itemize}

%


\section{Related Works}
\subsection{Visible-infrared Person Re-ID}
Visible-infrared person Re-ID handles person retrieval between different modalities, which is implemented by setting RGB cameras and IR cameras. Considering the poor illumination condition in some cases, especially at night, this cross-modal task does enjoy practical significance. Thanks to the image-based RGB-IR person Re-ID dataset constructed by Wu \textit{et al.} \cite{wu2017rgb}, many cross-modal Re-ID technologies have been studied, most of which are based on metric learning\cite{ye2018hierarchical, ye2018visible,  hao2019hsme, feng2019learning, ye2020cross, ye2019bi, li2020infrared, ling2020class },
feature learning\cite{ye2018hierarchical, ye2018visible, ye2019modality, ye2020cross, ye2019bi, ye2020dynamic, park2021learning}, and
adversarial learning\cite{dai2018cross, wang2019rgb, wang2019learning, choi2020hi, lu2020cross, choi2020hi, lu2020cross, pu2020dual}, etc.

As for metric learning and feature learning, Ye \textit{et al.} \cite{ye2018hierarchical} extracted multi-modal shareable features in the feature learning stage, after which heterogenous features are projected into a common space and measured by the metric learning.
Hao \textit{et al.} \cite{hao2019hsme} maps the extracted features onto a hypersphere manifold, in which differences
between two samples are calculated based on angles.
Different from angle measurement in \cite{hao2019hsme}, Feng \textit{et al.}\cite{feng2019learning} utilized the Euclidean constrain to shrink the cross-modal gap.
In \cite{ye2018visible}, inter-modality and intra-modality variations are firstly taken into account, and the discriminative features are then learned through a top-ranking loss.
Ye \textit{et al.} \cite{ye2019bi} then extended \cite{ye2018visible} by introducing a bi-directional center-constrained top-ranking loss, further improving the image based person Re-ID performance.

Additionally, the generative adversarial network (GAN) \cite{goodfellow2014generative} has also been widely applied to cross-modal Re-ID tasks.
Dai \textit{et al.} \cite{dai2018cross} embedded a discriminator into the network, enforcing the features from two modalities to be unclassified in an adversarial way.
In \cite{wang2019rgb}, by introducing the CycleGAN \cite{zhu2017unpaired}, an RGB image is transformed to an IR version, while its id-information is preserved. So is the IR image.
Furthermore, some researchers \cite{choi2020hi, lu2020cross} also achieved feature learning in an adversarial and disentanglement learning way.
Particularly, Choi \textit{et al.} \cite{choi2020hi} disentangled id-discriminative features and id-excluded features from cross-modal images, which are then combined to generate modal-different but id-consistent images. However, this strategy encounters a large computational complexity and some generated images with poor quality do give an inferior influence on the performance.

Apart from aforementioned methods, Ye \textit{et al.} \cite{ye2021channel} additionally focused on the image properties of RGB/ IR images. For instance, a novel joint learning strategy by channel augmentation and simulating random occlusions is proposed.
Moreover, image alignment \cite{park2021learning} and pattern alignment \cite{wu2021discover} are also exploited to alleviate the discrepancies.

\subsection{Video-based Person Re-ID}
Different from image-based person Re-ID, video-based person Re-ID represents a person by a sequence of images, providing temporal-information and a richer appearance\cite{ye2021deep}. Generally, existing methods mainly adopt RNN\cite{mclaughlin2016recurrent, xu2017jointly, liu2019spatial, zhang2017learning},
temporal pooling (average or weighted)\cite{xu2017jointly,chung2017two, wu2018exploit, gao2018revisiting},
optical flow\cite{mclaughlin2016recurrent, zhang2017learning,chen2018video},
and 3D convolution\cite{gu2020appearance, liu2019spatial, li2019multi}, etc.
For instance, Xu \textit{et al.} \cite{xu2017jointly} took original person images and corresponding optical flow as the network inputs, so that the motion consistency for a person in different periods is ensured. Then, features extracted from the CNN-RNN module are utilized to compute attention vectors, selecting informative frames over the sequence. A novel Spatial and
Temporal Memory Networks (STMN) \cite{eom2021video} is proposed, in which features for spatial distractors that frequently emerge across video frames, as well as attentions optimized for typical temporal patterns, are both stored.
Furthermore, Aich \textit{et al.}\cite{aich2021spatio} designed a flexible feature processing module which can be used in any 3D convolutional block for the Re-ID task. Thanks to this module, complementary person-specific appearance and motion information are well captured. Besides, 3D graph convolution is also introduced for video-based Re-ID. Liu \textit{et al.} \cite{liu2019spatial} employed context-reinforced topology to build a graph, which successfully encodes contextual information and physical information of the human body. By applying the 3D graph convolutional layers to it, spatial-temporal dependencies and structural information are efficiently captured.


Despite the fact that a number of works have been done for person ID, they are only either image-based cross-modal Re-ID or video-based RGB Re-ID.
Our HITSZ-VCM is the first dataset combining cross modalities
and video data, allowing the study on video-based cross-modal person Re-ID.
It not only achieves the 24-hour surveillance, but also gets comprehensive information and obtains much higher Re-ID accuracy.

%
%

\begin{table*}
	\centering
	\caption{Comparison between HITSZ-VCM with some typical Re-ID datasets.}
	\begin{tabular}{l|ccccccc}
		\toprule
		Dataset & Type & \#Identites & \#RGB cam. & \#IR cam. & \#Images \& BBoxes & \#Tracklets & Evaluation \\ \hline
		iLIDS-VID\cite{wang2014person}    & Video    & 300  & 2  & 0 & 42,495  &  600   &  CMC         \\
		MARs\cite{zheng2016mars}   &  Video  &  1,261 & 6   &  0  &  1,067,516  & 20,715   & CMC + mAP    \\
		Duke-Video\cite{wu2018exploit}& Video &  1,812 & 8 &   0 & 815,420  &  4,832  & CMC + mAP \\
		LS-VID\cite{li2019global}& Video &  3,772 & 15 &   0 & 2,982,685  &  14,943  & CMC + mAP \\ \hline
		RegDB\cite{nguyen2017person}& Image  & 412 & 1 &  1  & 8,240 &  -  & CMC + mAP \\
		SYSU-MM01\cite{wu2017rgb}&Image &  491 & 4 & 2 &  303,420  &  -  &  CMC + mAP \\ \hline
		HITSZ-VCM& Video & 927 & 12  &  12 & 463,259  & 21,863 & CMC + mAP \\
		\bottomrule
	\end{tabular}
	
	\label{tab:datasets}
	
\end{table*}

\section{Dataset}
\subsection{Dataset Description}
In this paper, we build the HITSZ-VCM (HITSZ Video Cross-Modal) Re-ID dataset for the video-based cross-modal person Re-ID task. To our best knowledge, this is the first cross-modal Re-ID dataset based on videos. The HITSZ-VCM dataset contains a large amount of images/frames captured by 12 HD cameras with 3840 $\times$ 2160 resolution. Thanks to the modern monitoring technology, all the cameras can automatically shoot both RGB and IR images according to the lighting conditions. Thus, each person is captured by both RGB and IR cameras. 
Note that all tracklets are processed by an automatic object tracking system, and we then finetune inaccurate annotations manually.


In detail, our HITSZ-VCM dataset contains 927 valid identities. The cameras shoot 25 frames per second and we extract the first frame out of every 5 frames to build the final dataset. According to this setting, every 24 consecutive images are regarded as a tracklet for a person during the same period, and the last frames whose number may be less than 24 form the last tracklet. Totally, there are 251,452 RGB images and 211,807 IR images, which can be divided into 11,785 and 10,078 tracklets, respectively. Of course, the number of frames in a tracklet can also be dynamically set, which is more flexible than many existing video-based datasets. More specifically, 12 cameras are used for our video collection. Generally, most of identities are captured by 3 RGB cameras and 3 IR cameras, and these cameras are non-overlapped.

\begin{figure}
	\centering
	\includegraphics[width=0.9\linewidth]{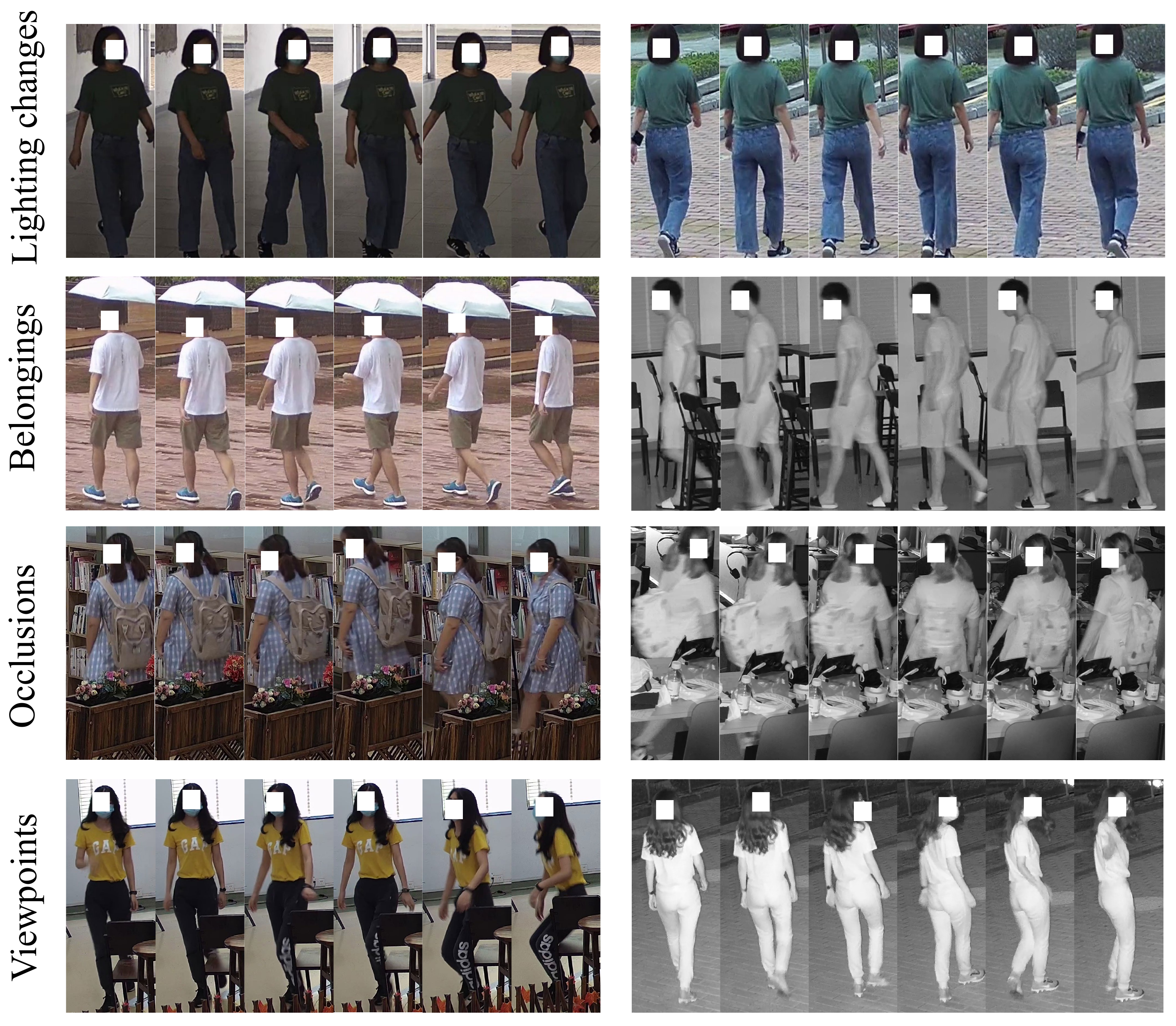}
	
	\caption{Some challenging tracklets in our dataset, including lighting changes, belonging changes, occlusions and viewpoint changes.}
	\label{fig:dataset show}
\end{figure}


Our HITSZ-VCM dataset also covers a series of diverse scenarios. Firstly, 7 outdoor, 3 indoor and 2 passages scenes are included. In detail, some common venues like the office, cafe, passageway, playground, and garden are all considered. Besides, each person is captured from multiple angles under each camera, constructing a richer appearance set. Furthermore, some challenging scenarios such as lighting changes (for RGB images), belonging changes, occlusion and viewpoint changes are collected, as displayed in \cref{fig:dataset show}.

\cref{tab:datasets} tabulates the comparison between HITSZ-VCM and existing related Re-ID datasets. As we can see, although MARs \cite{zheng2016mars}, Duke-Video \cite{wu2018exploit}, and LS-VID \cite{li2019global} also enjoy a large number of valid identities, they fail to cover IR images or videos, being incapable for the 24-hour surveillance. In contrast to RegDB \cite{nguyen2017person} and SYSU-MM01 \cite{wu2017rgb}, our constructed HITSZ-VCM dataset extends the image-based version to the video-based one, which provides more abundant and valuable information for person Re-ID. Furthermore, there are much more identities captured from more diverse scenarios in our dataset, greatly contributing to the training of the deep network.


In conclusion, HITSZ-VCM enjoys the following characteristics: (1) Constructing the first video-based cross-modal dataset for person Re-ID. (2) Collecting much more valid identities under diverse scenes. (3) Covering challenging but practical cases.

\subsection{Evaluation Protocol}
Here we conduct cross-camera and cross-modal retrieval like existing works \cite{wu2017rgb, zheng2016mars, li2019global,nguyen2017person}.
In other words, query and gallery are captured by different cameras and modalities.
Meanwhile, with respect to the `probe to gallery' pattern, video-to-video matching is adopted to keep consistent with training data.
Regularly, we utilize two retrieval modes for HITSZ-VCM: `infrared to visible' and `visible to infrared', to achieve a more comprehensive evaluation.
Additionally, we take all the tracklets in one modality as the query set and those from the other modality as the gallery set.
Totally, in the `infared to visible' retrieval, there respectively exist 5,159 and 5,643 tracklets in the query set and the gallery set. Vice versa in the `visible-to-infrared' mode.
Note that in our implementation, we discard some too short tracklets (less than 12 images).

To quantitatively evaluate the performance on our proposed dataset, the Cumulative Matching Characteristic curve (CMC) and mean Average Precision(mAP) are adopted as the evaluation metrics. 
Being similar to many methods, we compute the distance scores of all the query features and gallery features to do the ranking work. For testing, cosine similarity is used as the distance measurement.


\begin{figure*}
	\centering
	\includegraphics[width=0.85\linewidth]{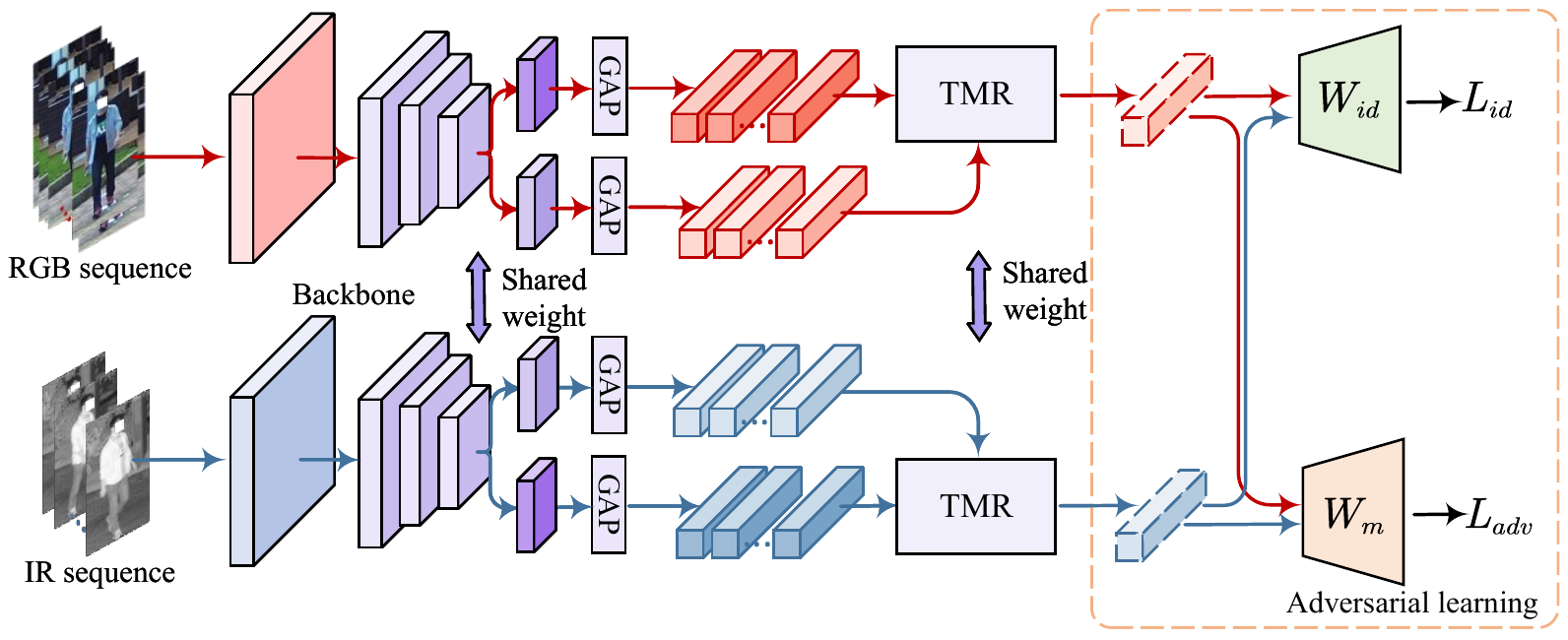}
	
	\caption{The framework of our proposed method. RGB/IR image sequences are regarded as cross-modal inputs. The Temporal Memory Refinement (TMR) module aggregates frame-level features into sequence-level features. $W_{id}$ and $W_m$ lead the classification of identities and modalities. By adopting an adversarial learning strategy, modal-related information is removed from the cross-modal data, and only id-related features are preserved.}
	\label{fig:framework}
\end{figure*}


\section{Baseline}
We follow the baseline proposed in \cite{ye2020dynamic}. A two-stream network, with ResNet50\cite{he2016deep} utilized as the backbone\cite{wang2019rgb, wang2019learning, ye2020dynamic}, is employed to handle the heterogeneous data belonging to different modalities.
Specifically, the first convolutional blocks in two branches enjoy different weights, so that modal-specific features for RGB and IR sequences are learned, respectively. Differently, in the remaining four blocks, the weights are shared to extracted modal-invariant features for these two modalities. Since the inputs of the network are multiple images, an average pooling layer is utilized to fuse the frame-level features obtained from the backbone. Thus, the sequence-level feature of each tracklet is finally obtained. By following \cite{ye2020dynamic}, the identity loss is introduced to guide the intra-modal Re-ID task, while the triplet loss is exploited to handle the cross-modal Re-ID task.
Therefore, the objective function $\mathcal{L}^{base}$ can be formulated as follows:
\begin{equation}\footnotesize
\mathcal{L}^{base} = \mathcal{L}_{id}^{base} + \mathcal{L}_{tri}^{base}
\end{equation}
where $\mathcal{L}_{id}^{base}$ and $\mathcal{L}_{tri}^{base}$ denote the identity loss and the triplet loss, respectively.

\section{Proposed Method}
Based on the baseline, a novel method is further proposed to more efficiently learn the modal-invariant and time-memory features for RGB and IR modalities.
The framework of our model is illustrated in \cref{fig:framework}.
By modifying the last convolution block in the backbone to two-branch convolution blocks (shared structures but different weights), two sets of feature maps from a sequence are obtained and then forwarded into a Temporal Memory Refinement (TMR) module, so that the temporal-information is extracted to meet the motion consistency for an identity. Furthermore, to fill the gap between two modalities, two classifiers are introduced, through which the modality-related features are removed while the id-related features are enhanced, greatly contributing to the cross-modal retrieval.

%
%

\subsection{Temporal Memory Refinement}
Here we respectively denote an RGB sequence and an IR sequence as
$\mathbf{V}=\left\{ \mathbf{V}^t\left| \mathbf{V}^t\in \mathbb{R} ^{H\times W} \right. \right\} _{t=1}^{T}$
and
$\mathbf{I}=\left\{ \mathbf{I}^t\left| \mathbf{I}^t\in \mathbb{R} ^{H\times W} \right. \right\} _{t=1}^{T}$, where $H$ and $W$ denote the height and weight of the images, $t$ means the $t$-th frame of this sequence, and $T$ is the total number of images in a tracklet.
Correspondingly, the ID labels are denoted as $p_v$ and $p_i$, while $m_v$ and $m_i$ denote the modal labels.

\begin{figure}
	\centering
	\includegraphics[width=0.87\linewidth]{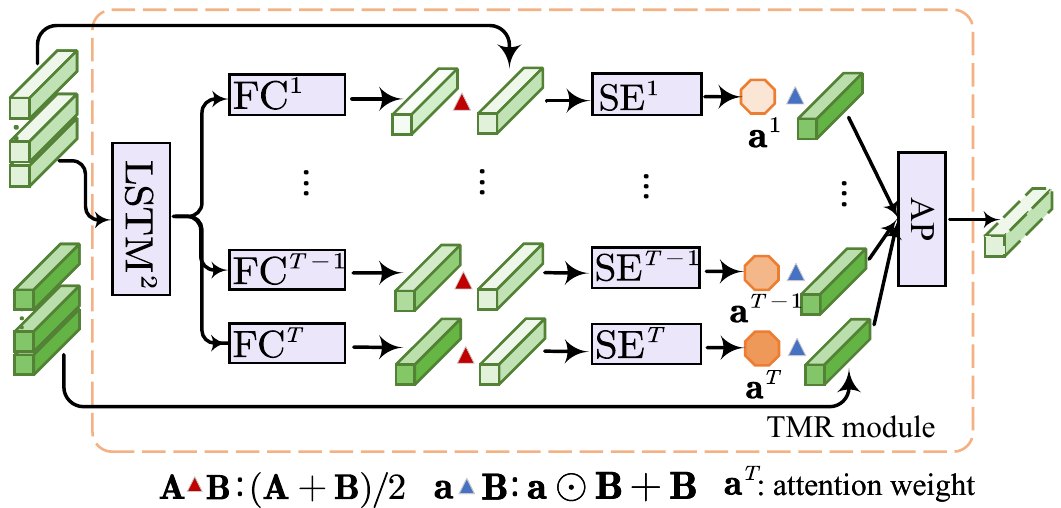}
	
	\caption{The insight of the TMR module. Multiple images from a tracklet are refined and the frame-level features are aggregated into a sequence-level one.}
	\label{fig:LSA}
\end{figure}

To transform frame-level features into a sequence-level feature and effectively capture temporal contexts among multiple frames, inspired by \cite{eom2021video}, we propose a Temporal Memory Refinement (TMR) module.
The structure of TMR is shown in \cref{fig:LSA}.
LSTM \cite{hochreiter1997long} layers and the SE attention \cite{hu2018squeeze} jointly facilitate the exploitation of temporal information, refining the features to enjoy more discriminative information.

Take RGB data $\mathbf{V}$ as an example and the five convolution blocks in our backbone are denoted as ${E}_{res}$.
We denote the two sets of frame-level features from ${E}_{res}$ as $\mathbf{f}_{v1}=\{\mathbf{f}_{v1}^{t}\}_{t=1}^{T}$ and $\mathbf{f}_{v2}=\{\mathbf{f}_{v2}^{t}\}_{t=1}^{T}$, which are utilized for the attention weights generation and the frame-level features aggregation, respectively.
Features $\mathbf{f}_{v1}$  are first forwarded into two LSTM layers $LSTM^{2}$, so that the temporal context of this tracklet is obtained. By applying a full-connected layer $FC^{t}$ to its associated output from $LSTM^{2}$ and adding $\mathbf{f}_{v1}^{t}$, an attention $\mathbf{a}^{t}$ is obtained by following the SE attention module.
\begin{equation}\small
\mathbf{a}^t = SE^{t}\left( \left( FC^t\left( LSTM^{2}(\mathbf{f}_{v1}) \right) +\mathbf{f}_{v1}^{t} \right) /2 \right) ,
\label{eq: se}
\end{equation}
where $SE^{t}$ means the $t$-th SE attention module. Note that $\mathbf{a}^t$ denotes the temporal-memory which gives the importance of the $t$-th frame in a tracklet. In other words, $\mathbf{a}^t$ plays as the attention weight of the $t$-th frame-level feature.

Based on the aforementioned analysis, the person representation of the $t$-th frame could be refined
and the average pooling processing is then utilized to aggregate the frame-level features into a sequence-level one:
\begin{equation}\footnotesize
\mathbf{F}_v=\sum_{t=1}^T{\left( \mathbf{a}^t\odot \mathbf{f}_{v2}^{t}+\mathbf{f}_{v2}^{t} \right)}/T
\label{eq: average pooling}
\end{equation}
Similar processing is conducted for IR data with the shared weights, through which its sequence-level $\mathbf{F}_{i}$ is obtained.

Overall, the temporal information is captured by the TMR module, so that the person representations are refined within a single modality. In our training phase, this module is optimized simultaneously with $E_{res}$, which is regarded as a supplementary for $E_{res}$.

\subsection{Modal-Invariant Learning}
After exploiting TMR where the intra-modal features are refined, we then introduce a modal-invariant adversarial learning to remove the gap from two modalities. Referring to the adversarial strategy, AlignGAN proposed in \cite{wang2019rgb} transforms the RGB/IR image to the IR/RGB version in the pixel-level by confusing a modal-discriminator. However, this strategy is constrained on the image generation, which not only increases the computational complexity, but also is quite sensitive to the quality of the generated image. By contrast, inspired by \cite{li2021dual}, here we achieve the adversarial learning only based on the feature-level, more efficiently getting the modal-invariant features.

%


According to Eq.(\ref{eq: average pooling}), the sequence-level features for RGB and IR tracklets are $\mathbf{F}_{v}$ and $\mathbf{F}_{i}$, respectively. Theoretically, if $\mathbf{F}_{v}$ and $\mathbf{F}_{i}$ do enjoy the id-related information but without modal-related information, they cannot be classified to $m_{v}$ or $m_{i}$. To achieve this task, a classifier $W_{m}$ is introduced whose output is a $3\times 1$ vector. Particularly, this output denotes the probabilities of a tracklet belonging to the RGB modality, IR modality, or neither of them. The objective function is formulated as:
\begin{equation}\small
\mathcal{L}_{adv1}\left( E \right) =CE\left( W_m\left( \mathbf{F}_v \right) , m_3 \right) +CE\left( W_m\left( \mathbf{F}_i \right) , m_3 \right)
\label{eq: adv1}
\end{equation}
where $E$ is the combination of the backbone $E_{res}$ and TMR, $CE(\cdot)$ denotes the cross-entropy loss, and $m_{3}$ means the third category which is neither belonging to the RGB modality $m_{v}$ nor the IR modality $m_{i}$. To encourage $\mathbf{F}_{v}$ and $\mathbf{F}_{i}$ to enjoy the discriminative information on identities, we also apply the id-related cross entropy loss and triplet loss to them via another classifier $W_{id}$. Thus, the id-related but modal-invariant function can be represented as:
\begin{equation}\small
\mathcal{L}_{id}\left( E, W_{id} \right) =\mathcal{L}_{adv1}+\mathcal{L}_{id}^{ce}+\mathcal{L}_{id}^{tri}
\label{eq: keep}
\end{equation}
where $\mathcal{L}_{id}^{ce}$ and $\mathcal{L}_{id}^{tri}$ are the id-related cross-entropy loss and triplet loss, respectively.

Of course, the classification capability of $W_{m}$ plays a key role for the modality-invariant feature learning. Here, being similar to existing GAN based methods, the adversarial learning strategy is adopted to additionally update $W_{m}$, shown as follows:
\begin{equation}\small
\mathcal{L}_{adv2}\left( W_m \right) =CE\left( W_m\left( \mathbf{F}_v \right) , m_v \right) +CE\left( W_m\left( \mathbf{F}_i \right) , m_i \right)
\label{eq: adv2}
\end{equation}

\begin{figure}
	\centering
	\includegraphics[width=0.9\linewidth]{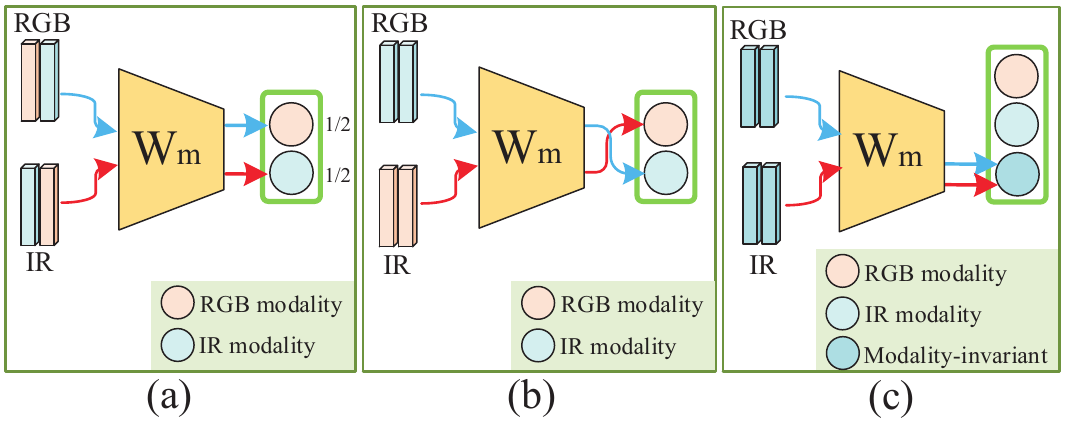}
	
	\caption{Different strategies in adversarial learning. (a) UCDA\cite{qi2019novel} and MCLNet\cite{hao2021cross}; (b) cmGAN\cite{dai2018cross}; (c) Ours.}
	\label{fig:gan}
\end{figure}

In the optimization, Eq.(\ref{eq: keep}) and Eq.(\ref{eq: adv2}) are optimized in an alternative way, so that $W_{m}$ enjoying the more strong capacity adversarially contributes to the modal-invariant feature learning. Note that, it is true that some work \cite{qi2019novel, hao2021cross, dai2018cross} also learn the id-related features via the adversarial learning in the feature level. However, they are different from each other. As displayed in Fig.\ref{fig:gan}(a), UCDA\cite{qi2019novel} and MCLNet\cite{hao2021cross} enforce the probability belonging to each category to be the same. 
MCLNet deceives the network to confound different modalities while UCDA acts camera-aware domain adaptation. However, a feature which contains both RGB and IR information can also have the same classification result.
Obviously, in this case, we are unsure that whether these two modalities are aligned. Referring to cmGAN\cite{dai2018cross} in Fig.\ref{fig:gan}(b), RGB and IR features are inversely classified to IR and RGB modalities. A limitation is that these two inputs are transformed to only gain each other modal-specific information, while the modal-gap is not measured. By contrast, our used strategy directly classifies different features into an additional class, guaranteeing that they do fall in a same latent space or domain.

\begin{table*}[t]\small
	\centering
	\caption{Effectiveness of the TMR module and the adversarial learning module. Note that M denotes the modal-invariant learning, T denotes the temporal memory refinement(TMR), and Full method$^{S}$ denotes full method with shuffled frames in TMR.}
	\begin{tabular}{l|ccccc|ccccc}
		\toprule[0.8pt]
		\multirow{2}{*}{Strategy} & \multicolumn{5}{c|}{\textit{Infrared to Visible}} & \multicolumn{5}{c}{\textit{Visible to Infrared}} \\ \cline{2-11}
		& R1    & R5    & R10    & R20    & mAP    & R1    & R5    & R10    & R20    & mAP   \\ \hline
		Baseline  & 55.58   & 70.75  & 77.01  & 82.16     & 40.80  & 59.58    & 74.43   & 79.25    & 83.74    & 42.61    \\
		Baseline + M & 59.73   &   74.50  &   80.06     &   84.69  &  42.80  & 60.54   &   75.84    &   81.15   &   85.59    &  43.59        \\	
		Baseline + T &  58.44  &  72.32   &  78.51    &   83.51     &  43.87  &  62.19   &  76.11   &  80.84   &   84.98    &  46.02     \\
		Full method$^{S}$ &   60.82    &   74.54     &  78.69  &   83.25    &   43.94 &   63.72  &   77.72  &  82.70   &  86.90     &  45.64     \\
		Full method  &   \textbf{63.74}    &   \textbf{76.88}     &  \textbf{81.72}    &   \textbf{86.28}     &  \textbf{ 45.31 }   &   \textbf{64.54}    &   \textbf{78.98}   &  \textbf{82.98}  &   \textbf{87.10}   &  \textbf{47.69}     \\
		\bottomrule[0.8pt]
	\end{tabular}
	
	\label{tab: ablation}
\end{table*}

\begin{table*}[]\small
	\centering
	\caption{Comparisons of our modal-invariant learning with different adversarial strategies shown in \cref{fig:gan}. For a fair comparison, we only replace our modal-invariant learning with other strategies in our network.}
	\begin{tabular}{l|ccccc|ccccc}
		\toprule[0.8pt]
		\multirow{2}{*}{Strategy} & \multicolumn{5}{c|}{\textit{Infrared to Visible}} & \multicolumn{5}{c}{\textit{Visible to Infrared}} \\ \cline{2-11}
		& R1    & R5    & R10    & R20    & mAP    & R1    & R5    & R10    & R20    & mAP   \\ \hline
		cmGAN\cite{dai2018cross} &  57.96     &   72.58    &    78.32     &  83.42     &   43.14    &  60.68     &   75.25    &    80.29     &  85.00     &   44.63       \\
		UCDA\cite{qi2019novel} &  59.51      &   73.34    &    79.14     &    84.18       &    45.06      &  64.15  &   77.47    &    82.35    &   86.21  &   47.08      \\	
		Our method &   \textbf{63.74}   &   \textbf{76.88}    &  \textbf{81.72 }    &   \textbf{86.28  }    &  \textbf{ 45.31 } &  \textbf{64.54} & \textbf{78.96 }& \textbf{82.98} &\textbf{87.10} & \textbf{47.69}    \\
		\bottomrule[0.8pt]
	\end{tabular}
	
	\label{tab: adv}
\end{table*}

\section{Experiment}

\subsection{Experimental Setting }
\noindent \textbf{Dataset.} Here we divide our dataset into two sets for training and testing. The training set contains 500 identities with 232,496 images and 11,061 tracklets, while the testing set contains 427 identities with 230,763 images and 10,802 tracklets.
In the training phase, all the images are resized to a size of $288\times144$. Being similar to many existing methods, random cropping with zero-padding and horizontal flipping are also used for the data augmentation.


\noindent \textbf{Experimental Implementation.} We implement our model by PyTorch\cite{paszke2019pytorch} and train it on a NVIDIA TESLA A100 with cuda version 11.2.
The ResNet50\cite{he2016deep} pretrained on the ImageNet\cite{deng2009imagenet} is exploited our baseline and backbone.
For the encoder $E$ and id-classifier $W_{id}$, they are optimized via the optimizer SGD with the weight decay of $5\times 10^{-4}$ and the momentum of 0.9.
We adopt a learning rate warmup strategy for $E$ and $W_{id}$ and its initial value is set to 0.1.
After 35 epochs and 80 epochs, the learning rate is reduced to 0.01 and 0.001, respectively.
Note that the learning rate of the first unshared convolution blocks is always the one tenth of that of the remaining modules.
In terms of the modal-classifier $W_{m}$, the SGD optimizer works with the 0.01 learning rate, the $5\times 10^{-4}$ weight decay and 0.9 momentum.
We set the maximum number of epochs to 200.
Besides, the batch size is set to 16 with 8 different identities and 2 tracklets for each identity.

Furthermore, for each tracklet with 24 continuous images, $n$ (it can be dynamically selected) images are selected for training.
Specifically, 24 images are divided into $n$ parts with $24/n$ images, in which 1 of $24/n$ images is randomly selected to form the training data. 
Besides, to synchronize RGB and IR tracklets, we shuffle all tracklets first and then select the same number of RGB and IR tracklets batch by batch. When all IR tracklets are selected, we shuffle all tracklets again.

\begin{figure}
	\centering
	\includegraphics[width=0.8\linewidth]{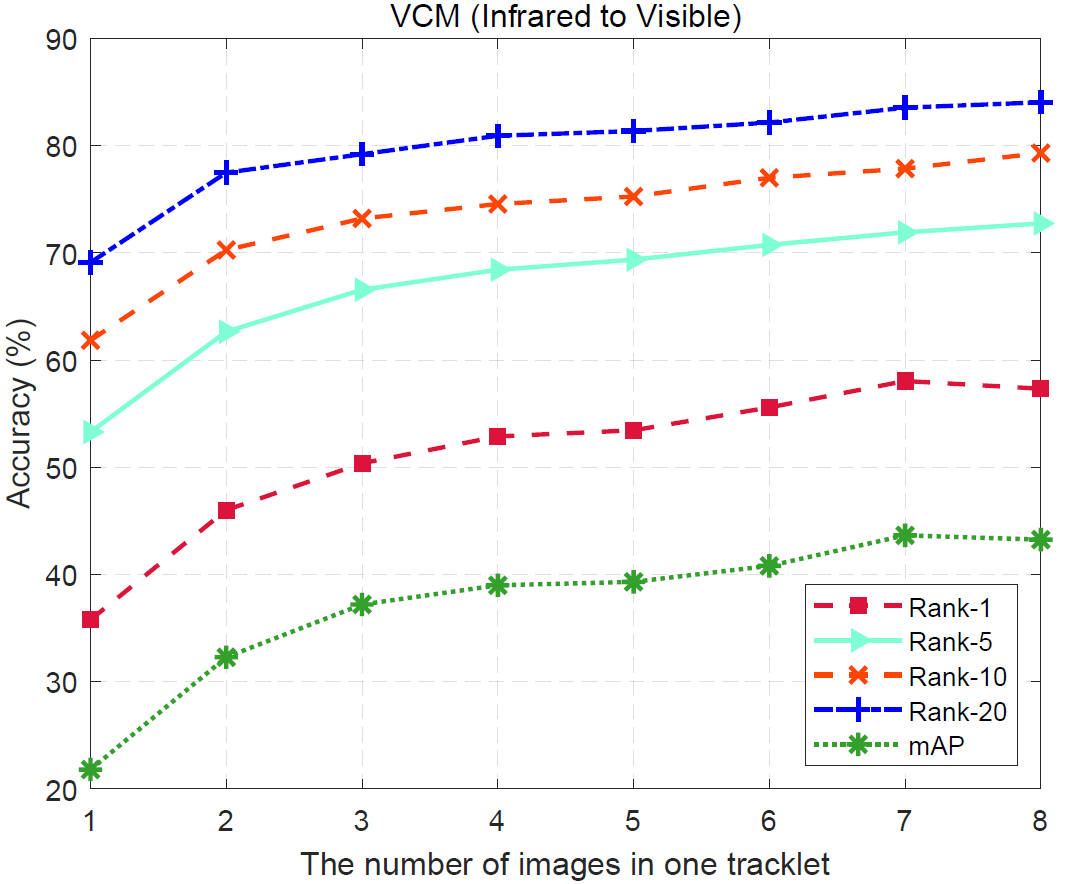}
	
	\caption{Evaluation of our baseline on different settings, where ${n}$ denotes the number of images in one tracklet.}
	\label{fig:baseline_seq}
\end{figure}

\begin{figure}
	\centering
	\includegraphics[width=0.88\linewidth]{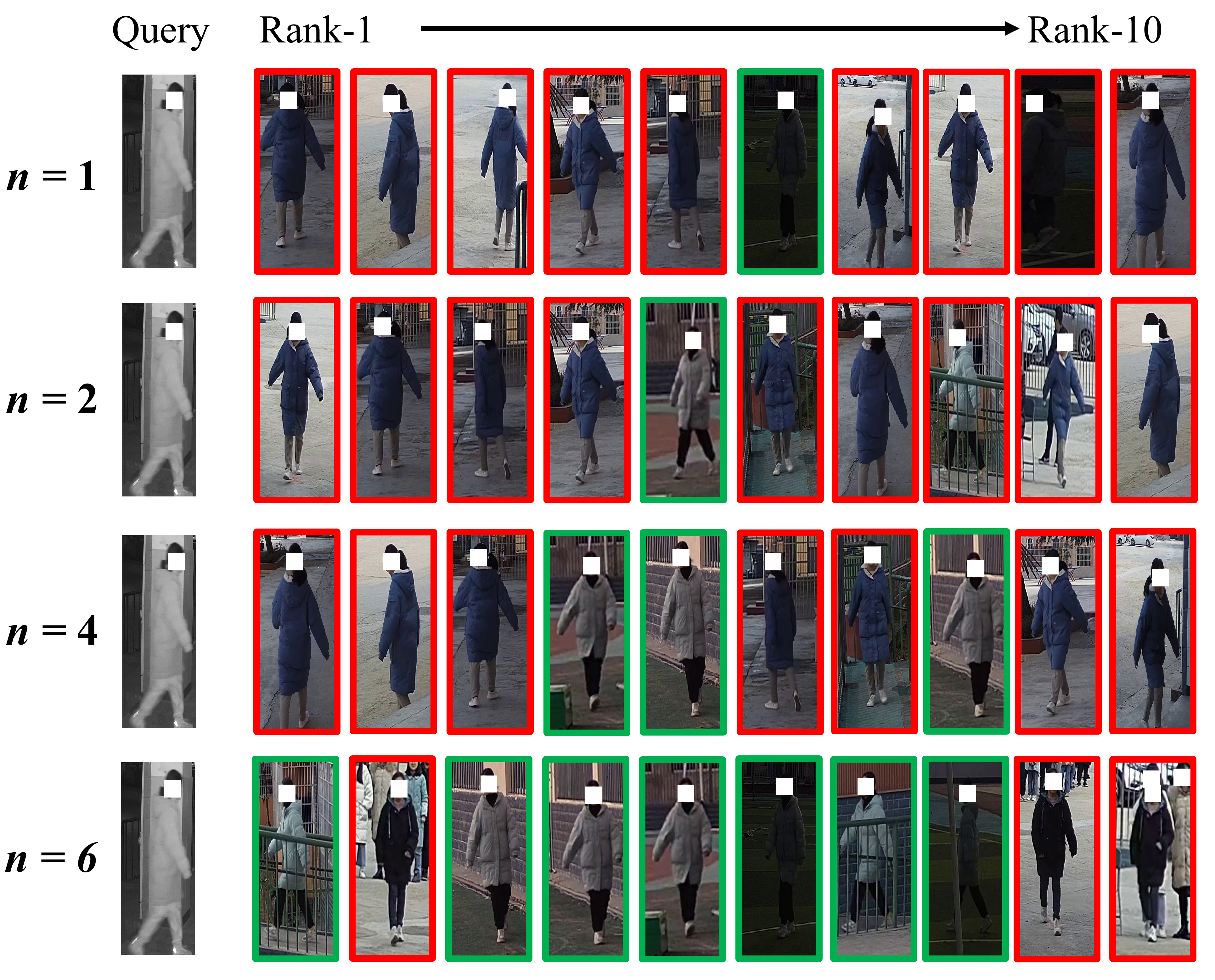}
	
	\caption{The visualization results on different settings, where ${n}$ denotes the number of images in one tracklets. Obviously, image-based methods (${n}=1$) cannot report satisfying results when two identities enjoy similar appearances, while video-based methods show noticeable performance with the enhancement of temporal information.}
	\label{fig:case}
\end{figure}

\subsection{Ablation Study}
\label{sec: ablation}
In this subsection, we experimentally analyze the significance of our HITSZ-VCM dataset, as well as the strategies in the proposed method MITML.

\begin{table*}[]\small
	\centering
	\caption{Comparisons of our method with state-of-the-art cross-modal methods on our HITSZ-VCM dataset. CMC (\%) and mAP (\%) are reported.}
	\begin{tabular}{l|c|ccccc|ccccc}
		\toprule[0.8pt]
		\multirow{2}{*}{Method} & \multirow{2}{*}{Venue} & \multicolumn{5}{c|}{\textit{Infrared to Visible}} & \multicolumn{5}{c}{\textit{Visible to Infrared}} \\ \cline{3-12}
		&       & R1    & R5    & R10    & R20    & mAP    & R1    & R5    & R10    & R20    & mAP   \\ \hline
		LbA\cite{park2021learning}     & ICCV'21                & 46.38     & 65.29     & 72.23      & 79.41      & 30.69      & 49.30  & 69.27  & 75.90   & 82.21     & 32.38     \\
		MPANet\cite{wu2021discover}    & CVPR'21                & 46.51     & 63.07  & 70.51   & 77.77   & 35.26   & 50.32 & 67.31  & 73.56   & 79.66   & 37.80  \\	
		DDAG\cite{ye2020dynamic}       & ECCV'20                & 54.62     & 69.79     & 76.05      & 81.50      & 39.26      & 59.03    & 74.64     & 79.53      & 84.04      & 41.50     \\
		VSD\cite{tian2021farewell}    & CVPR'21                &54.53  & 70.01 & 76.28   & 82.01& 41.18  & 57.52 & 73.66 & 79.38  & 83.61  & 43.45 \\
		CAJL\cite{ye2021channel}      & ICCV'21                & 56.59   & 73.49   & 79.52    & 84.05   & 41.49   & 60.13  & 74.62  & 79.86   & 84.53   & 42.81   \\\hline
		Ours    & -     &\textbf{63.74} & \textbf{76.88}  & \textbf{81.72} & \textbf{86.28 } & \textbf{45.31}  & \textbf{64.54} & \textbf{78.96 }& \textbf{82.98} &\textbf{87.10} & \textbf{47.69} \\
		\bottomrule[0.8pt]
	\end{tabular}
	
	\label{tab:comparison}
\end{table*}

\noindent \textbf{Significance of video-based cross-modal datasets.}
Compared with image data, video data provide more abundant information for the person Re-ID task.
To verify the above statement, we conduct the experiments by changing the number of images in one tracklet on our baseline, as displayed in \cref{fig:baseline_seq}.
It can be seen that with the increase of images in a tracklet, the Re-ID performance meet a continuous rise, demonstrating the significance of our constructed dataset. Specifically, when only one image is exploited, like that in existing image-based cross-modal Re-ID dataset, the mAP is only 23.09\%, which is much inferior to that when six images are simultaneously used in a tracklet. \cref{fig:case} further illustrates the Top-10 visualization results of our proposed method MITML with different settings on datasets, also substantiating the necessity of our HITSZ-VCM dataset.



As shown in \cref{fig:baseline_seq}, the values of all metrics increase really slightly when $n$ is relatively large, i.e., 7 and 8. To reduce the time costs in the training phase, we set $n$ to 6 in the following experiments.

\noindent \textbf{Effectiveness of temporal-information exploitation.}
As shown in \cref{tab: ablation}, we evaluate the performances of TMR in our approaches on our dataset.
Compared with `baseline', our TMR module (`Baseline + T' in \cref{tab: ablation}) achieves a remarkable performance improvement on rank-1 and mAP, respectively.
The main reason is that the temporal-information extracted by TMR facilitates to build an excellent appearance model and obtain person unique features which cannot be captured from image-based data.
Also, we shuffle frames in a tracklet (`Full method$^{S}$'in \cref{tab: ablation}), and the results are inferior, indicating the importance of temporal information.

\noindent \textbf{Effectiveness of modal-invariant learning.}
The adversarial learning used in MITML successfully removes the modal-related information from different modalities but also preserves id-related features. As tabulated in \cref{tab: ablation}, by adding this module into the baseline (`baseline + M'), there is indeed a performance enhancement, which also substantiates its effectiveness.

Furthermore, as shown in \cref{tab: adv}, we evaluate the other two adversarial learning strategies as discussed in \cref{fig:gan}. As we can see, the learning strategy in MITML is more effective than that in \cite{qi2019novel} and \cite{dai2018cross}.

\noindent \textbf{Loss function parameter adjustment.} We also evaluate loss functions with different weighted terms. Consider that the term $\mathcal{L}_{adv1}$  to a great extent determines the learning of modal-invariant features, here we mainly test the weight of this term. Thus, the Eq.(\ref{eq: keep}) can be represented as: 
\begin{equation}\small
\mathcal{L}_{id}\left( E, W_{id} \right) =\lambda\mathcal{L}_{adv1}+\mathcal{L}_{id}^{ce}+\mathcal{L}_{id}^{tri}
\label{eq: lambda}
\end{equation}
where $\lambda$ is the weight of term $\mathcal{L}_{adv1}$. As shown in \cref{tab: weight}, as $\lambda$ changes from 0.2 to 1.0, the experimental results fluctuate slightly, which indicates that our algorithm is robust.
When $\lambda$ is 0.4, our proposed model achieves best performance.

\begin{table}[]
	\centering
	\caption{Evaluations of loss function with different weight.}
	\begin{tabular}{c|ccccc}
		\toprule[0.8pt]
		\multicolumn{6}{c}{\textit{Infrared to Visible}}\\\hline
		Parameter $\lambda$     & \multicolumn{1}{c}{R1} & \multicolumn{1}{c}{R5} & \multicolumn{1}{c}{R10} & \multicolumn{1}{c}{R20} & \multicolumn{1}{c}{mAP} \\ \hline
		0.01     & 60.80   & 74.76     & 80.06   & 85.17    & 45.54       \\
		0.05   & 63.05   &   76.81   &   81.83    &86.08     &  46.50    \\
		0.1 &  62.98    &  76.40    &  81.15   &  85.54  &  46.19     \\
		0.2 &   63.53  &   76.81      &  81.26   &  85.89    &  46.49             \\
		0.4 &   \textbf{64.62}  &   \textbf{77.12}      &  \textbf{82.37}   &  \textbf{86.81}    &  \textbf{47.32}  \\
		0.6 &   63.46  &   76.83      &  81.65   &  85.71    &  45.71             \\
		0.8 &   63.48  &   76.83      &  82.07   &  86.34  &  45.46             \\
		1.0 &   63.74  &   76.88  &  81.72   &  86.28 &  45.31  \\
		\bottomrule[0.8pt]
	\end{tabular}
	
	\label{tab: weight}
\end{table}

\noindent \textbf{Baseline pooling strategy.} Since the inputs of the
network are multiple images, an average pooling layer is
utilized in our baseline to fuse the frame-level features obtained from the
backbone. Here, we conduct different pooling strategies for baseline, including max pooling and weighted average pooling, as shown in \cref{tab: pooling}.
Note that `weighted pooling' in \cref{tab: pooling} denotes weighted average pooling, and we compute the attention scores based on the Softmax function.
As we can see, the max pooling strategy is also adaptive for our baseline.

\begin{table}[]\small
	\centering
	\caption{ Evaluations of different pooling strategies in baseline. `Weighted pooling' means weighted average pooling, the attention scores of which are computed based on Softmax function.}
	\begin{tabular}{l|ccccc}
		\toprule[0.8pt]
		\multicolumn{6}{c}{\textit{Infrared to Visible }}\\\hline
		Strategy     & \multicolumn{1}{c}{R1} & \multicolumn{1}{c}{R5} & \multicolumn{1}{c}{R10} & \multicolumn{1}{c}{R20} & \multicolumn{1}{c}{mAP} \\ \hline
		Average pooling &  55.58  &  70.75   &   77.01    &  82.16 &   40.80   \\
		Max pooling &  54.47  & 70.24  &  76.51  &  81.83   &   41.11    \\
		Weighted pooling &   47.49 & 64.88& 72.43 &79.04 & 36.52    \\
		\bottomrule[0.8pt]
	\end{tabular}
	
	\label{tab: pooling}
\end{table}

\subsection{Comparison with State-of-the-art Methods}
In this section, we further compare our proposed method with existing state-of-the-art visible-infrared cross-modal person Re-ID methods,
including DDAG\cite{ye2020dynamic}, LbA\cite{park2021learning}, MPANet\cite{wu2021discover} and VSD\cite{tian2021farewell} and CAJL\cite{ye2021channel}. Note that, these comparison methods are primarily designed for image-based datasets. For a fair comparison, we conduct an average pooling layer for their generated frame-level features. For those networks whose backbones are ResNet50, including \cite{ye2020dynamic, park2021learning, wu2021discover, tian2021farewell}, we implement the average pooling after the backbone, which is similar to what we did in our baseline. Furthermore, we pretrain the model for CAJL \cite{ye2021channel} on AGW \cite{ye2021deep} before the channel augmented joint learning.

%
%
%

The CMC and mAP obtained by all these approaches on our dataset are listed in \cref{tab:comparison}. Obviously, our method reports a noticeable improvement than these state-of-the-art imaged-based cross-modal approaches.
Specifically, for the infrared to visible retrieval mode, Rank-1 and mAP increase by 7.15\% and 3.82\% respectively than the second best method CAJL.
As for the visible to infrared search mode, Rank-1 and mAP also obtain a significant growth of 4.41\% and 4.88\%,  indicating the effectiveness of our TMR module on the temporal information exploitation.

\vspace{-5pt}

\subsection{Limitation}
Based on the above analysis, the importance  of video-based cross-modal person Re-ID is proved, and  our methods demonstrates a more remarkable improvement compared with existing methods.
However, our methods requires a fixed number of images in one tracklet in the training and testing phases, which decreases the flexibility in realistic applications.
In our future work, we will aim to design a novel network which can process tracklets with dynamic lengths.

\vspace{-3pt}

\section{Conclusion}
Based on the observation that video data can provide temporal-information and allow us to build a richer appearance model for identification,
we study a new task: video-based cross-modal person Re-ID.
To achieve this goal, the first video-based cross-modal Re-ID dataset is constructed.
There exist 927 valid identities with 251,452 RGB images of 11,785 tracklets and 211,807 IR images of 10,078 tracklets captured by 12 HD RGB/IR cameras, in which 500 identities for training and 427 identities for testing. Experimental results prove the significance of our constructed dataset.
Additionally, a novel method: modal-invariant and temporal memory learning (MITML) is proposed for our HITSZ-VCM dataset.
Specifically, an adversarial learning strategy contributes to extracting the high-quality modal-invariant features and bridging the modal heterogeneity,
while a temporal memory refinement module effectively captures the motion consistency.
Although video-based cross-modal person Re-ID is a challenging task, our proposed method achieves remarkable performance, compared with existing state-of-the-art cross-modal approaches.

\section*{Acknowledgement}
This work was supported in part by Shenzhen Science and Technology Program (RCBS20200714114910193) and the NSFC fund (61906162, 61966021).

\newpage

{\small
	\bibliographystyle{ieee_fullname}
	\bibliography{egbib}
}

\end{document}